%% file: Latex-template.tex
\documentclass[A4paper, 11 pt, conference]{ieeeconf}  
\IEEEoverridecommandlockouts       
\usepackage{geometry}
\usepackage{cite}

\usepackage{graphics} 
\usepackage{epsfig} 
\usepackage{mathptmx} 
\usepackage{times} 
\usepackage{amsmath} 
\usepackage{amssymb}  
\usepackage{subfigure}
\usepackage{algorithm} 
\usepackage{algorithmic} 
\usepackage{multirow} 
\usepackage[english]{babel}
\usepackage{verbatim} 

\begin{document}

\title{\LARGE \bf
A Q-learning Control Method for a Soft Robotic Arm Utilizing Training Data from a Rough Simulator}

\author{Peijin Li$^1$, Gaotian Wang$^2$, Hao Jiang$^1$, Yusong Jin$^1$, Yinghao Gan$^1$, Xiaoping Chen$^1$, and Jianmin Ji$^1$ 

}

\maketitle 
\thispagestyle{empty}

\begin{abstract}
It is challenging to control a soft robot, where reinforcement learning methods have been applied with promising results. 
However, due to the poor sample efficiency, reinforcement learning methods require a large collection of training data, which limits their applications. 
In this paper, we propose a Q-learning controller for a physical soft robot, in which pre-trained models using data from a rough simulator are applied to improve the performance of the controller. 
We implement the method on our soft robot, i.e., Honeycomb Pneumatic Network (HPN) arm. 
The experiments show that the usage of pre-trained models can not only reduce the amount of the real-world training data, but also greatly improve its accuracy and convergence rate.
\end{abstract}

\let\thefootnote\relax\footnotetext{
$^*$This research is supported by the Open Project Fund from Shenzhen Institute of Artificial Intelligence and Robotics for Society, under Grant No. AC01202005010 and the National Natural Science Foundation
of China under grant number 61573333. The work is partially supported by  CAAI-Huawei MindSpore Open Fund, and Key-Area Research and Development Program of Guangdong Province 2020B0909050001.

$^1$Peijin Li,  Hao Jiang, Yusong Jin, Yinghao Gan, Xiaoping Chen, and Jianmin Ji are with the School of
Computer Science, University of Science and Technology of China, Hefei,
230026, China. Jianmin Ji is the corresponding author{\tt\small jianmin@ustc.edu.cn}.

$^2$Gaotian Wang is with the school of physical Science, University of Science and Technology of China, Hefei, 230026, China.

}

\input{sections/introduction}

\input{sections/HPN_arm}

\input{sections/method}
\input{sections/experiment}
\input{sections/conclusion}

\bibliography{ref.bib}
\bibliographystyle{IEEEtran}

\end{document}

%% file: sections/introduction.tex
\section{Introduction}

Soft robots possess infinite degrees of freedom and inherent compliance, which expand their potential applications such as medical care~\cite{deng2013development}, domestic service~\cite{jiang2021hierarchical}, and performing various tasks in confined spaces~\cite{marchese2016design}.
However, it is typically challenging to generate a realistic mathematical model for a soft robot due to its complex structure and nonlinear features. 
Thus it is hard to control a soft robot following classical control methods. 

On the other hand, machine learning methods have shown remarkable potentials in handling control problems for various soft robots.
These methods can generally be classified into two categories: model-based methods and model-free methods. 
A model-based method usually first learns a model of the soft robot and controls the robot using the learned model. 
For instance, Giorelli et al~\cite{giorelli2015neural} use a feed-forward neural network to learn the static model of a cable-driven soft robot and achieve position control of its tip. 
They also prove that their neural network-based controller is better than a Jacobian model-based static controller in terms of accuracy and convergence rate, in the sense that the former can theoretically consider all kinds of uncertainties. 
Melingui et al~\cite{melingui2014qualitative} learn an inverse kinematics model of a soft robot based on distal supervised learning and implement a controller. 
In their next work~\cite{melingui2015adaptive}, they add adaptive sub-controllers to deal with the effects of hysteresis and viscoelasticity. 
Thuruthel et al~\cite{thuruthel2016learning} use a neural network to learn the differential inverse kinematics of a soft robot.
In their next work~\cite{george2017learning}, they add the pose of the tip of the robot to the input of the neural network to deal with external disturbances. 

A model-free method usually follows a model-free reinforcement learning method to learn a control policy directly.
You et al~\cite{you2017model} use a Q-learning method to learn a controller for a 2-D soft robotic arm. 
Ansari et al~\cite{ansari2017multiobjective} use cooperative multi-agent reinforcement learning to control the stiffness and position simultaneously. 
However, due to the poor sample efficiency of reinforcement learning, their methods require a large collection of training data, which limits their applications.

Recently, there have been some works that try to reduce the amount of the required training data. 
Jiang et al~\cite{jiang2021hierarchical} propose a method that generates data by setting virtual goals to reduce the need for data from real arms and implement a Q-learning based controller that can be trained in real-time. 
However, the state defined in the Q-learning does not include the information of the pose of the goal and the tip of the arm in the absolute coordinate system, which limits the ability of the controller to adjust the control policy according to this information. 
Thus, the performance of this controller to some goals is not expected to be acceptable.
In fact, the need for real-world training data can be reduced by making use of training data in simulation. 
Satheeshbabu et. al~\cite{satheeshbabu2019open} use a model proposed by Uppalapati et. al~\cite{uppalapati2018design} to train a DQN (Deep Q-Network) based controller, and use state feedback to apply it to a physical soft robot, which avoids the cost of collecting data from the real-world. 
However, the state definition of the DQN does not include the information of the pose of the goal and the tip of the arm in the absolute coordinate system either. 
When there is more than one goal, the performance is not expected to be good.
In their next work~\cite{satheeshbabu2020continuous}, the actuation is added to the state, which includes the information of the absolute pose of the tip and the goal in theory.
However, when the soft robot interacts with environments or there are large external loads, the information of the absolute pose of the tip indicated by the actuation may be incorrect.

In this work, we propose a Q-learning controller for a physical soft robot that uses pre-trained models by utilizing training data from a rough simulator for the robot. 
We implement the method on our soft robot, i.e., Honeycomb Pneumatic Network (HPN) arm. The HPN structure is first introduced in~\cite{sun2014towards}.
As shown in Figure\ref{hard_ware_pic}, a soft robot, i.e., HPN arm, is provided in~\cite{jiang2016design}.
The experiments show that our method is robust and the usage of pre-trained models can not only reduce the amount of the real-world training data, but also greatly improve its accuracy and convergence rate.

%% file: sections/HPN_arm.tex

\begin{figure}[htp]
    \centering
    \includegraphics[width=0.5\columnwidth]{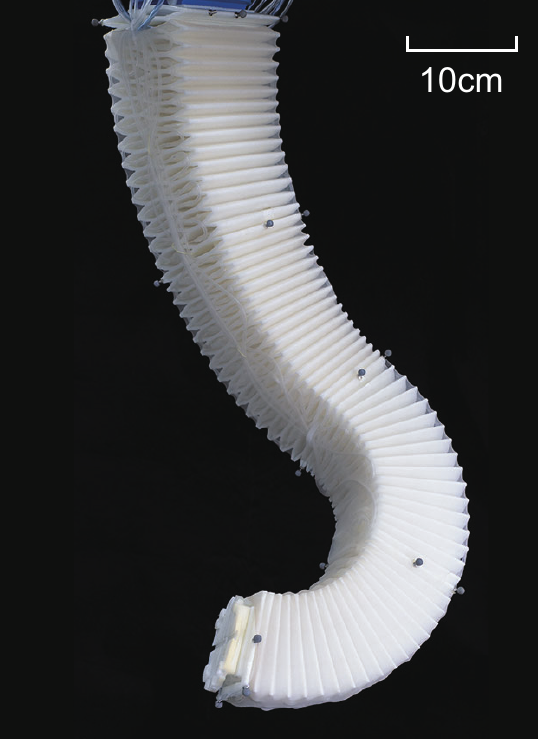}
    \caption{The HPN arm with four segments.  
There are markers on the base and the tip of the arm so that the pose of the tip with respect to the base can be obtained by the motion capture system (MCS, Prime 17W, OptiTrack). 
Each segment is composed of a deformable Honeycomb structure and four groups of airbags, which can be pressurized independently.
By pressurizing the four groups of airbags with different pressure, the HPN structure can deform differently.
}
    \label{hard_ware_pic}
\end{figure}

%% file: sections/method.tex
\section{Methods}
We specify our Q-learning control method in this section.
We first introduce the model of the simulator. 
Then we provide the Q-learning method and the pre-training method based on the simulator.

\subsection{The PCC model of the simulator.}

\begin{figure}[htp]
    \centering
    \includegraphics[width=0.6\columnwidth]{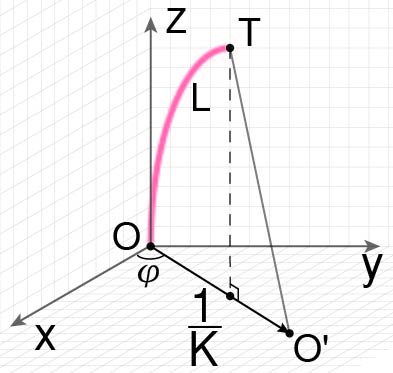}
    \caption{The illustration of the parameters of the configuration space. The red curve represents the central axis of a single segment of the HPN arm. We establish a coordinate system at the centroid of the base of the arm $O$, let the positive direction of the $z$-axis be the tangent direction of the central axis at the base of the arm, $O'$ is the center of the arm's curvature, $O O'$ is the radius, which is equal to the reciprocal number of the curvature 
   $\frac{1}{K}$, and $L$ is the arc length of the arm, $\varphi$ is the angle between the plane of the arm and the $x$-$z$ plane.}
    \label{configuration_space_pic}
\end{figure}

The simulator is based on the assumption of piece-wise constant curvature (PCC)~\cite{webster2010design}.
Note that, PCC models are widely adopted for controlling various soft robots. 
A PCC model is consisted of two mapping: 
(1) the mapping from configuration space to task space; 
(2) the mapping from the actuation space to the configuration space. 
The former mapping has a general solution, while the latter is arm-specific. 
We use the pressure of four groups of airbags of each segment as the actuation space parameter, and use three parameters, $K$, $\varphi$, and $L$ as the configuration space parameter, which represent the curvature, the plane in which the arm bends, and the arc length respectively, as shown in Figure~\ref{configuration_space_pic}.


We use a homogeneous transformation matrix that transforms the base to the tip of the whole arm as the task space parameter. 
In the following, we specify the two mappings. 


\subsubsection{Mapping from the actuation space to the configuration space}

The mapping from the actuation space to the configuration space is to transform the pressure of four groups of the airbags of a single-segment HPN arm into the three parameters in the configuration space: $K$, $\varphi$, and $L$. 

\begin{figure}[htbp]
    \centering
    \includegraphics[width=0.3\columnwidth]{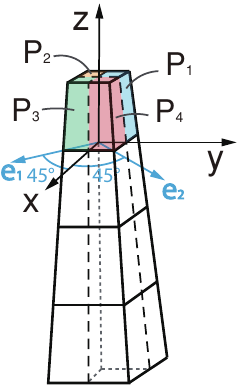}
    \caption{The arrangement of the airbags of the arm. The four prisms represent four segments of the arm. We take the first segment as an example. $P_1$, $P_2$, $P_3$, and $P_4$ are the pressure of four groups of airbags in the direction shown in the figure. Vector $\mathbf{e_1}$ and vector $\mathbf{e_2}$ are the unit vectors whose angle with the positive direction of the $x$-axis are $45^{\circ}$.}
    \label{actuation_pic}
\end{figure}

As shown in Figure~\ref{actuation_pic}, for a single-segment HPN, we define a coordinate system in which the centroid of the base of the arm is the origin and the positive direction of the $z$-axis is toward the tangential direction of the central axis at the base of the arm. 
Then the plane where the arm bends in is always perpendicular to the $x$-$y$ plane. 
Thus, the bending direction of the arm can be represented by $\varphi$, which is the angle between the plane of the arm and the $x$-$z$ plane. 


Figure~\ref{actuation_pic} illustrates the arrangement of the airbags of a single segment of the arm. 
Pressurizing a certain group of airbags can make the arm bend towards the opposite side of the airbags. 
As shown in Figure~\ref{actuation_pic}, $\mathbf{e_1}=[\frac{\sqrt{2}}{2},-\frac{\sqrt{2}}{2}]$, $\mathbf{e_2}=[\frac{\sqrt{2}}{2},\frac{\sqrt{2}}{2}]$, they are the unit vectors whose angle between the positive direction of the $x$-axis are $45^{\circ}$. 
According to the working principle of the HPN arm, we assume that pressurizing $P_1$ makes the arm bend towards the direction of $\mathbf{e_1}$, pressurizing $P_3$ makes the arm bend towards the direction of $-\mathbf{e_1}$, pressurizing $P_2$ makes the arm bend towards the direction of $\mathbf{e_2}$, and pressurizing $P_4$ makes the arm bend towards the direction of $-\mathbf{e_2}$. 
So the bending of the arm can be resolved into two directions: $\mathbf{e_1}$ and $\mathbf{e_2}$. 

We assume that the bending direction of the single-segment arm is the linear combination of the two bending directions mentioned above. 
The direction vector representing the bending direction of the arm can be expressed as $\mathbf d= \frac{(P_{1}-P_{3})\mathbf{e_{1}}+(P_{2}-P_{4})\mathbf{e_2}}{\left|(P_{1}-P_{3})\mathbf{e_{1}}+(P_{2}-P_{4})\mathbf{e_2}\right|}$, where  $(P_1-P_3) \mathbf{e_1}$ is the vector representing the bending along the direction of $\mathbf{e_1}$, $(P_2-P_4)\mathbf{e_2}$ is the vector representing the bending along the direction of $\mathbf{e_2}$. 

Furthermore, the angle between the plane in which the arm is and the $x$-$z$ plane can be represented as $\varphi= \arcsin( \mathbf d \cdot \mathbf x)$.  
$\mathbf x=[1,0]$ is the unit vector in the direction of the positive $x$-axis. 
In order to simulate the feature of HPN that the greater the difference of the pressure is, the greater the bending will be, we assume that the curvature of the HPN arm is proportional to the linear combination of the bending in two directions. 
Specifically, the curvature $K=A\left|(P_{1}-P_{3})\mathbf{e_{1}}+(P_{2}-P_{4})\mathbf{e_2}\right|$, where $A$ is a coefficient, which is set empirically here. 
Generally, the larger the sum of the pressure is, the longer the arm would be. 
In order to simulate this feature, we assume that the arc length of the arm is proportional to the sum of the pressures of all groups of airbags. 
Specifically, $L=B(P_1+P_2+P_3+P_4)+L_0$, where $L_0$ is the length of the arm with zero pressure in the airbags, and $B$ is a coefficient, which is also set empirically here.

\subsubsection{Mapping from the configuration space to the task space}

After obtaining the configuration space parameters, we need to calculate the task space parameters, i.e., the position and orientation of the tip of the whole arm according to the configuration space parameters of all segments. 



For each segment of the HPN arm, we represent the position and orientation of the tip by a homogeneous transformation matrix, 
\begin{equation}
T=\tiny{\left\{
\begin{matrix}
\cos^2 \phi (\cos K L-1)+1 & \sin \phi \cos \phi (\cos K L -1) & \cos \phi \sin K L & \frac{\cos\phi (1-\cos K L) }{K} \\ 
\sin \phi \cos \phi (\cos K L -1) & \cos^2 \phi (1-\cos K L ) + \cos K L & \sin \phi \sin K L & \frac{\sin\phi (1-\cos K L) }{K} \\ 

-\cos \phi \sin K L & -\sin \phi \sin K L & \cos K L &\frac{ \sin K L}{K} \\
0 & 0 & 0 & 1

\end{matrix}
\right\}}
\label{equation_T}
\end{equation}
for whom there is a general solution (equation \ref{equation_T}) \cite{webster2010design}.

After obtaining the homogeneous transformation matrix of all segments, the task space parameter can be obtained by multiplying the homogeneous transformation matrix of each segment in turn. Specifically, $T=T^1 T^2 T^3 T^4$, in which $T^1$, $T^2$, $T^3$, $T^4$ are the homogeneous transformation matrices of the first, second, third, and fourth segments respectively.


In order to test the accuracy of this model, we randomly selected 100 sets of pressure and compared the simulation results with the real actuating results. The average positional error of the simulation model is about 12cm, while the average rotational error is about $10^{\circ}$.


\subsection{Pre-training method and Q-learning controller.}


\subsubsection{The Q-learning controller}

Our Q-learning controller extends from the one in~\cite{jiang2021hierarchical} by specifying a new state definition and without adopting the method to generate training data by setting virtual goals. 
We first introduce the definition of state, action, and reward, and then introduce the complete Q-learning method. 

\begin{figure}[htbp]
    \centering
    \includegraphics[width=\columnwidth]{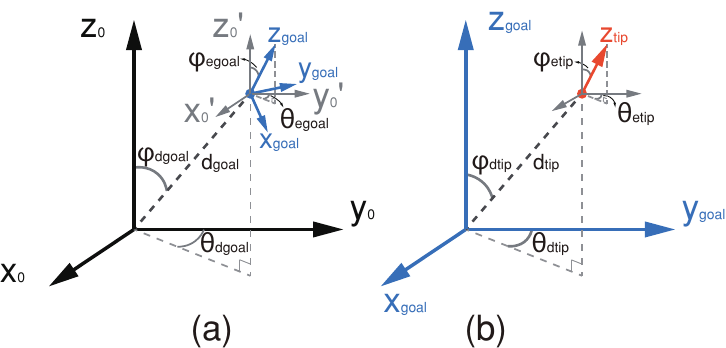}
    \caption{The illustration of the state definition.
     (a) The absolute pose of the goal which is represented by five parameters: $d_{goal}$, $\theta_{dgoal}$,$\varphi_{dgoal}$, $\theta_{egoal}$, $\varphi_{egoal}$. The first three parameters are the radius, azimuthal angle, and elevation angle of the goal, respectively. The last two are the azimuthal angle and the elevation angle of the orientation of the goal. (b) The pose of the tip with respect to the goal which is represented with five parameters: $d_{tip}$, $\theta_{dtip}$, $\varphi_{dtip}$, $\theta_{etip}$, $\varphi_{etip}$. The first three parameters are the radius, azimuthal angle, and elevation angle of the displacement vector of the tip with respect to the goal. The last two are the azimuthal angle and the elevation angle of the orientation of the tip with respect to the goal.
}
    \label{state_pic}
\end{figure}

The state definition consists of two parts: the absolute pose of the goal and the pose of the tip with respect to the goal.
As our HPN arm does not have the degree of freedom (DOF) of rotation, we do not control the rotation of the arm. 
Thus the two parts of the state can both be represented by a 5-D vector, which can be combined to a 10-D vector:
\small{$
(d_{goal}, \theta_{dgoal},\varphi_{dgoal}, \theta_{egoal}, \varphi_{egoal}, d_{tip}, \theta_{dtip}, \varphi_{dtip}, \theta_{etip}, \varphi_{etip})
$}, as shown in Figure~\ref{state_pic}~(a). 
The first five dimensions represent the absolute pose of the goal.
We choose the center of the workspace of the HPN arm as the origin of the spherical coordinate system in which we represent the absolute pose of the goal. 
In practice, we use the position of the tip when the arm is not actuated to approximate the center of the workspace. 
$d_{goal}$, $\theta_{dgoal}$, $\varphi_{dgoal}$ are the radius, azimuthal angle, and elevation angle of the goal respectively.  
$\theta_{egoal}$ and $\varphi_{egoal}$ are the azimuthal angle and the elevation angle of the orientation of the goal. 

As shown in Figure~\ref{state_pic}~(b), the last five dimensions represent the pose of the tip with respect to the goal. 
$d_{tip}$, $\theta_{dtip}$, $\varphi_{dtip}$ are the radius, azimuthal angle, and elevation angle of the displacement vector of the tip with respect to the goal. 
$\theta_{etip}$ and $\varphi_{etip}$ are the azimuthal angle and the elevation angle of the orientation of the tip with respect to the goal. 

We discretize the continuous state space. 
Specifically, the range of $d_{tip}$ is divided into four parts: $[0 mm, 5 mm)$, $[5 mm, 30 mm)$, $[30 mm, 60 mm)$, $[60 mm, +\infty)$, the range of $\varphi_{egoal}$ is divided into four parts: $[0^{\circ}, 5^{\circ})$, $[5^{\circ}, 20^{\circ})$, $[20^{\circ}, 60^{\circ})$, $[60^{\circ}, 180^{\circ}]$. 
The other eight parameters are divided into four parts evenly. 
Thus there are $4^{10}=1048576$ states in total.

The definition of the action and reward $^1$\footnote{$^1$Due to page limit, please refer to \cite{jiang2021hierarchical} for more details.} is the same as the Q-learning controller in~\cite{jiang2021hierarchical}.


In the process of Q-learning, we first pressurize the initial pressure to the HPN arm. Then in each turn of iteration, we first obtain the pose of the tip, and calculate the current state of the arm, choose an action according to the Q-table and $\epsilon$-greedy, perform the action, calculate the reward, and update Q-function according to the following equation.


{\small
\begin{equation*}
    Q(S_t,A_t)\leftarrow\alpha[R_{t}+\gamma \max\limits_{a}Q(S_{t+1},a)-Q(S_t,A_t)]+Q(S_t,A_t)
\end{equation*}}%
 where $S_t$, $A_t$ are the state and action of the $t$-th iteration, $R_t$ is the reward of that action.

\subsubsection{The pre-training method}

We randomly choose goals in the workspace of the HPN arm and use the Q-learning controller to control the HPN arm in the simulation to move to the goals to train the Q-table. Using simple motor babbling methods to choose training goals will result in the concentration of the training goals in the center of the workspace. In order to make the distribution of the goals more even, we choose goals in the task space directly. Each goal can determine the first five dimensions of the state, which represent the absolute pose of the goal. In order to make the distribution of the goals more even, we choose the same number of goals in each area determined by the first five dimensions of the state. 


Because the parts of the Q-table corresponding to the goals in different parts of the task space are independent of each other, parallel computation can be used to speed up the pre-training process. 

In order to further reduce the amount of data required for training, we augment the data in the Q-table trained with the above-mentioned goals in the simulation. Since our Q-table is very large, there may still be a large number of elements in the Q-table that have not been trained but maintain the initial value after the Q-table is trained with a large amount of data. We assume that the Q-values of the same actions in nearby states will be similar, so the average of the Q-values of its nearby states is used to replace these untrained initial values. 

%% file: sections/experiment.tex
\section{Experiment}
In this section, we evaluate our method by a series of experiments with the HPN arm, including the point-to-point experiment, and the comparing experiment with pre-trained models using the real HPN arm.


\subsection{Point-to-point experiment in free space.}

\begin{figure*}[htbp]
    \centering
    \includegraphics[width=\textwidth]{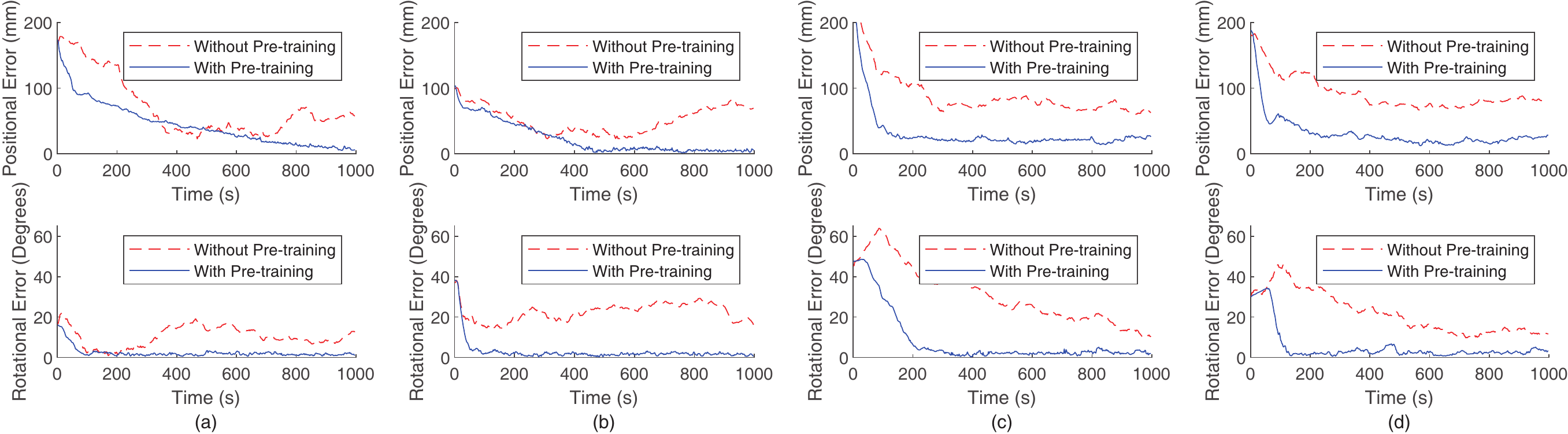}
    \caption{The result of the point-to-point experiment. The experiments of four goals are shown in the figure. (a) and (b) show the results of the two goals where the arm mainly needs to elongate, (c) and (d) show the results of the two goals where the arm mainly needs to bend. The red dashed curves show the results of the controller without pre-training, and the blue solid curves show the results of the controller with pre-training.  
}
    \label{exp1_pic}
\end{figure*}

In this experiment, we select four goals in the workspace of the HPN arm. 
In order to fully demonstrate the performance of our control method to reach different goals, we selected two goals that mainly need the arm to elongate and two goals that mainly need the arm to bend. Then we use our control method to control the arm to reach these goals, and compare the result with that of the control method without pre-training (with an initial Q-table whose elements are all zero). During the pre-training, we use the control method to reach 3072000 goals in the simulation to collect training data and train the Q-table. During the tasks, we recorded the positional and rotational error every two seconds. Each goal was reached three times by each method, and the positional and rotational errors of the three times were averaged to avoid the effect of stochastic factors.


The results show that our method can control the arm to reach the goal properly, and the pre-trained models can improve the performance of the controller greatly. Figure~\ref{exp1_pic} demonstrates the results. Figure~\ref{exp1_pic} (a) and (b) show the result of the two goals that mainly need the arm to elongate, Figure~\ref{exp1_pic} (c) and (d) show the result of the two goals that mainly need the arm to bend. 

It can be observed from Figure~\ref{exp1_pic} that the pre-training can greatly improve the accuracy and convergence rate of the control method. Figure~\ref{exp1_pic} (a) and (b) show that, when the goal mainly requires the arm to elongate, the convergence rate of the method without pre-training is slightly slower or almost the same as the method with pre-training. But without pre-training, after the positional error is at its minimum, it will slightly rise. This may be because the arm may move to states where it has never been to so that the controller does not have the proper strategy. As for the goals that mainly need the arm to bend, because of the effect of gravity, the PCC assumption which the model is based on does not fit the arm very well. So for these goals, the effect of pre-training is slightly worse. But it can be observed from Figure~\ref{exp1_pic} (c) and (d) that pre-training can still greatly improve the convergence speed and accuracy. Taking Figure~\ref{exp1_pic} (c) as an example, when there is pre-training, the position error can decrease to 23mm at 200 seconds. But when there is no pre-training, the position error can only decrease to 65mm at 300 seconds. It can be observed from Figure~\ref{exp1_pic} that for all four goals, the performance of rotational error has been greatly improved with pre-training. Specifically, when there is pre-training, the rotational error decreases slightly faster, and can remain at a lower level after it reaches its minimum. When there is no pre-training, the rotational error decreases slower, the accuracy is lower, and the rotational error may rise again after it reaches its minimum.

\subsection{The comparing experiment with the pre-training using physical HPN arm.}

Here we perform an experiment to compare the controller with the pre-trained model in simulation and the controller that is pre-trained using data collected on the physical HPN arm. 
We use the goal corresponding to Figure~\ref{exp1_pic} (a) in the point-to-point experiment as an example. We use the physical HPN arm to reach the goal a lot of times, and use the collected data to train the Q-table, which is used as the initial Q-table of the testing experiment. We found that the Q-table obtained after the point-to-point task for training was performed four times (4000 seconds) on the physical HPN arm can achieve almost the same performance as the Q-table pre-trained in the simulation. During the experiment, we recorded the positional and rotational error every two seconds. The goal was reached three times by each method, and the positional and rotational errors of the three times were averaged to avoid the effect of stochastic factors.

\begin{figure}[htbp]
    \centering
    \includegraphics[width=\columnwidth]{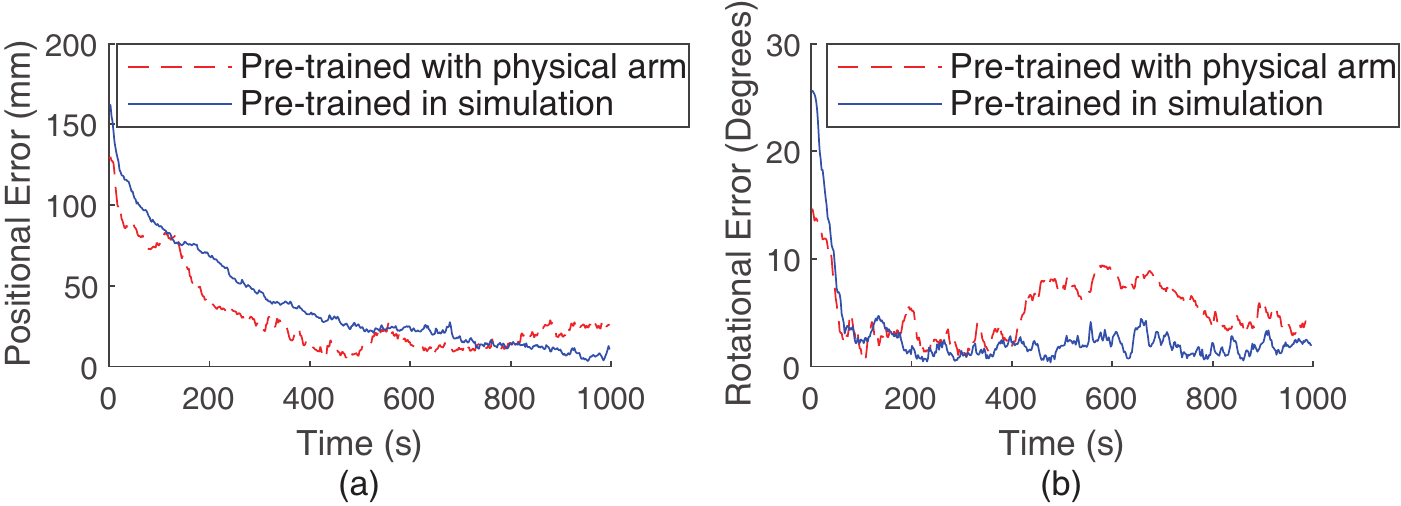}
    \caption{
    The comparing experiment with the pre-training using the physical HPN arm. The red dashed curve is the pre-training result with the physical HPN arm (four times of point-to-point task). The blue solid curve is the pre-training result with simulation. Figure (a) demonstrates the positional error during the testing experiment. Figure (b) demonstrates the rotational error during the testing experiment. 
}
    \label{exp2_pic}
\end{figure}

The results are demonstrated in Figure \ref{exp2_pic}. It can be observed from Figure \ref{exp2_pic} that the performance of the pre-training method proposed in this work is similar to the controller pre-trained with data collected using the physical HPN arm for 4000 seconds. Besides, the controller pre-trained with the physical HPN arm is only trained for a single goal. In order to train every part of the Q-table, at least 1024 times training data is required, which will cost 4096000 seconds to collect training data. While the pre-training method proposed in this work only takes about 614,400 seconds to reach 3072000 goals in simulation for pre-training without parallel computation (each goal takes about 0.2 seconds), which saves 3481600 seconds (about 967 hours), i.e., 85\% of the time. And with parallel computation, more time can be saved.

%% file: sections/conclusion.tex
\section{Conclusion}

In this paper, we propose a Q-learning controller for a physical soft robot arm, where pre-trained models using training data from a piecewise constant curvature (PCC) model based simulator are applied to improve the performance of the controller. 
We implement the method on our soft robot, i.e., Honeycomb Pneumatic Network (HPN) arm.  We have verified through experiments that this method can use simulation data to greatly improve the convergence accuracy and convergence rate of the Q-learning controller.
Moreover, we have verified through experiments that this method can greatly save the time and cost of collecting data. 